\PassOptionsToPackage{hidelinks,unicode}{hyperref}
\documentclass[sigconf]{acmart}

\usepackage{xcolor}

\usepackage{amsthm}
\usepackage{array}

\usepackage{xspace}
\AtBeginDocument{\RequirePackage{hyperxmp}}
\AtBeginDocument{\RequirePackage{cleveref}}

\usepackage{algorithm}
\usepackage{algpseudocode}
\usepackage{graphicx}
\usepackage{multirow}
\usepackage{subcaption}

\usepackage{amsthm}
\usepackage{xspace}
\theoremstyle{definition}

\usepackage{algorithm}
\usepackage{algpseudocode}
\usepackage{graphicx}

\usepackage{array}
\usepackage{multirow}
\usepackage{subcaption}
\usepackage{xcolor}
\usepackage{color}
\definecolor{black}{rgb}{0,0,0}
\definecolor{grey}{rgb}{0.8,0.8,0.8}
\definecolor{red}{rgb}{1,0,0}
\definecolor{green}{rgb}{0,1,0}
\definecolor{darkgreen}{rgb}{0,0.5,0}
\definecolor{darkpurple}{rgb}{0.5,0,0.5}
\definecolor{darkdarkpurple}{rgb}{0.3,0,0.3}
\definecolor{blue}{rgb}{0,0,1}
\definecolor{shadegreen}{rgb}{0.95,1,0.95}
\definecolor{shadeblue}{rgb}{0.95,0.95,1}
\definecolor{shadered}{rgb}{1,0.85,0.85}
\definecolor{shadegrey}{rgb}{0.85,0.85,0.85}
\definecolor{oddRowGrey}{rgb}{0.80,0.80,0.80}
\definecolor{evenRowGrey}{rgb}{0.85,0.85,0.85}

\definecolor{ForestGreen}{rgb}{0.0, 0.66, 0.47}
\definecolor{RubineRed}{rgb}{1.0, 0.0, 0.31}

\AtBeginDocument{%
  \providecommand\BibTeX{{%
    \normalfont B\kern-0.5em{\scshape i\kern-0.25em b}\kern-0.8em\TeX}}}

\setcopyright{none}

\begin{document}

\title[]{Classifier Language Models: Unifying Sparse Finetuning and Adaptive Tokenization for Specialized Classification Tasks}

\author{Adit Krishnan}
\affiliation{%
  \institution{Amazon Web Services Inc.}
  \city{Seattle}
  \state{WA}
  \country{USA}
}
\email{aditkris@amazon.com}

\author{Chu Wang}
\affiliation{%
  \institution{Amazon Web Services Inc.}
  \city{Seattle}
  \state{WA}
  \country{USA}
}
\email{chuwang@amazon.com}

\author{Chris Kong}

\affiliation{%
  \institution{Amazon Web Services Inc.}
  \city{Seattle}
  \state{WA}
  \country{USA}
}
\email{luyankon@amazon.com}
\begin{abstract}

Semantic text classification requires the understanding of the contextual significance of specific tokens rather than surface-level patterns or keywords (as in rule-based or statistical text classification), making large language models (LLMs) well-suited for this task. However, semantic classification applications in industry, like customer intent detection or semantic role labeling, tend to be highly specialized. They require annotation by domain experts in contrast to general-purpose corpora for pretraining. Further, they typically require high inference throughputs which limits the model size from latency and cost perspectives. Thus, for a range of specialized classification tasks, the preferred solution is to develop customized classifiers by finetuning smaller language models (e.g., mini-encoders, small language models). 

    In this work, we develop a token-driven sparse finetuning strategy to adapt small language models to specialized classification tasks. We identify and finetune a small sensitive subset of model parameters via task-specific token groups in the finetuning dataset (case study in \cref{appendix:tokens}), while leaving most of the pretrained weights unchanged. Unlike adapter approaches such as low rank adaptation (LoRA), we do not introduce additional parameters to the model. Our approach outperforms end-to-end finetuning, LoRA, layer selection, and prefix tuning on five diverse semantic classification tasks. We achieve greater stability, accuracy, and half the training costs compared to end-to-end finetuning.


\end{abstract}



\maketitle

\section{Introduction}
\label{sec:introduction}

\begin{figure}[t]
\centering  \includegraphics[width=0.42\textwidth]{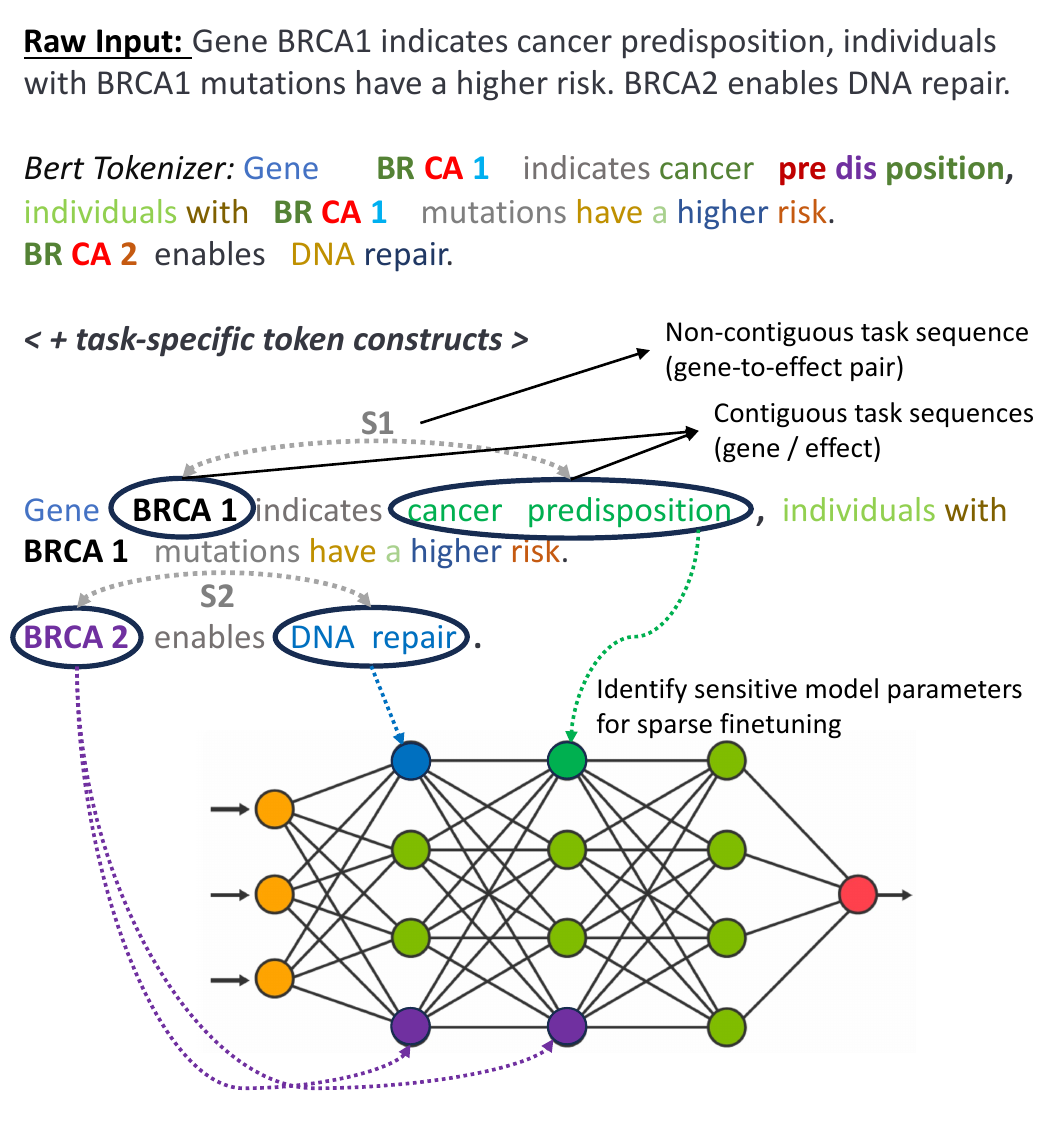}
  \vspace{-17pt}
  \caption{Task-specific tokenization helps finetune sensitive model parameters, while reusing most pretraining weights.}
  \label{fig:intro1}
  \vspace{-20pt}
\end{figure}

Cutting edge large language models (LLMs) such as Anthropic Claude-3\footnote{https://www.anthropic.com/claude} and OpenAI GPT-4~\footnote{https://openai.com/index/gpt-4/} dominate general-purpose language understanding benchmarks\footnote{https://paperswithcode.com/sota/multi-task-language-understanding-on-mmlu}. Although these models show impressive language understanding~\cite{mmlu}, they are not optimized for specialized semantic classification tasks that require domain expertise~\cite{nips2024llm}; for example, semantic role labeling in news content\footnote{https://paperswithcode.com/sota/semantic-role-labeling-on-ontonotes} where the best results belong to smaller task-specific models. In addition, a large-scale application typically serves several hundred to a thousand classification queries per second~\cite{scale1}, which is prohibitively expensive for large models. Thus, there has been an increasing focus on developing specialized, cost-effective language models for repetitive large-scale semantic classification tasks.

\begin{figure*}[]
  \centering
  \caption{We provide an illustration of prior sparse finetuning approaches compared to our proposed approach. The highlighted parameters (in orange) indicate the trainable components under each method.}
\includegraphics[width=0.9\textwidth]{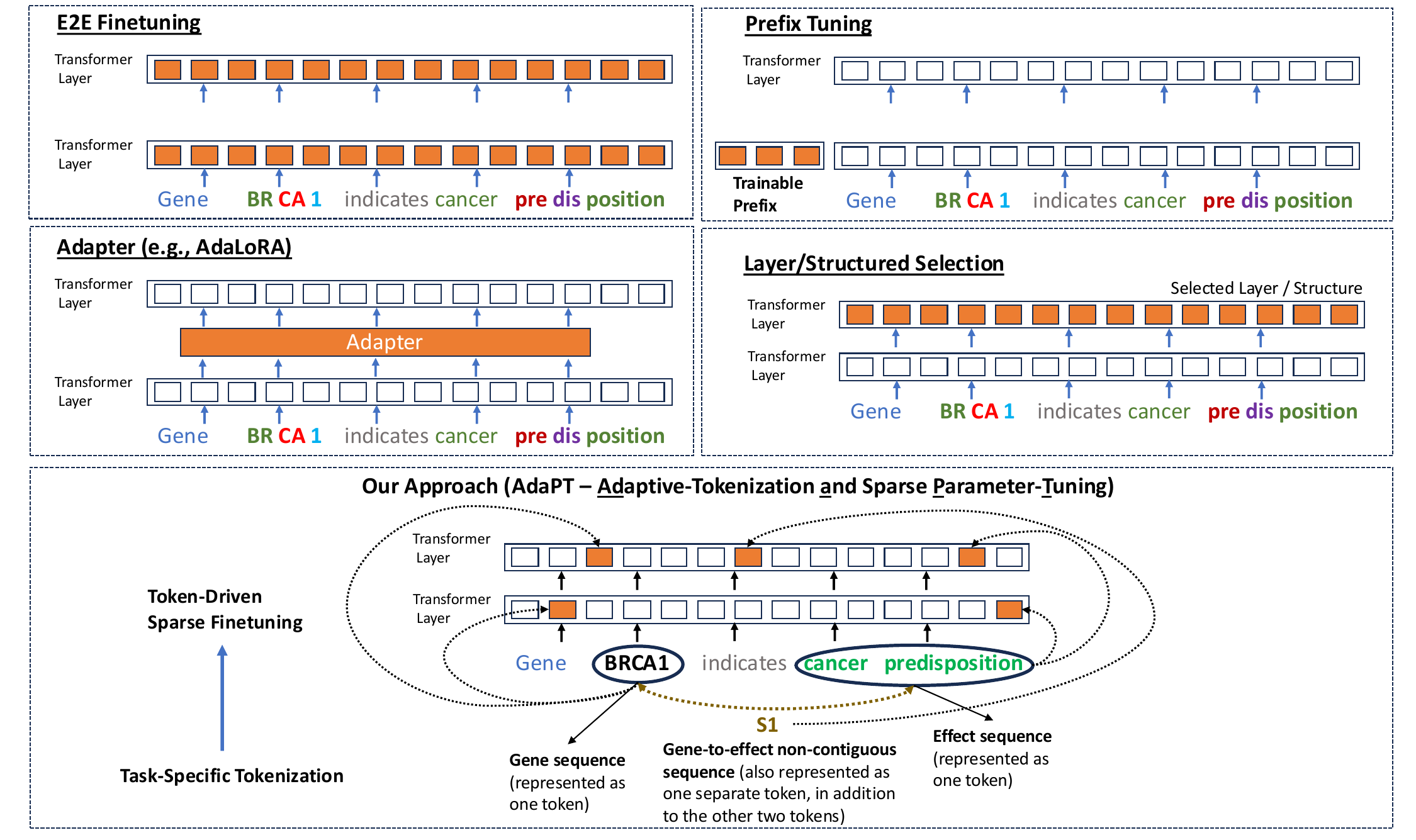}
  \label{fig:intro2}
  \vspace{-12pt}
\end{figure*}
Within the specialized model arena, finetuning small pretrained encoders (e.g., BERT and its variants) as well as small language models / decoders~\cite{phi2, gemma} has proven successful, both in performance~\cite{sun2019fine} and cost-efficiency~\cite{tsai2019small}. However, end-to-end (E2E) finetuning loses pretraining knowledge and is subject to overfitting risks and instability~\cite{mosbach2020stability}, especially with smaller finetuning datasets. The class of \textit{sparse finetuning methods} is designed to modify only a focused subset of the model parameters~\cite{neft, lester2021power, li2021prefix} for a new domain or task, thus mitigating the above risks. However, compared to E2E finetuning, they show performance limitations when the domain gap is large~\cite{liu2022improved}. In this paper, we answer the following question.

Can we combine the \emph{flexiblity and accuracy} of end-to-end finetuning with the \emph{stability and efficiency} of sparse finetuning for accurate and efficient semantic classification?

We identify token-driven parameter selection as the bridge to achieve task adaptation while retaining pretraining knowledge and the stability advantages of sparse finetuning. Our key insight is the following. By introducing task-specific phrases, token constructs, and subsequences to the model vocabulary, we can effectively estimate the sensitivity of each model parameter on the task-specific finetuning data. We then selectively finetune sensitive parameters for task adaptation while leaving most of the pretrained weights unchanged, thus reducing the risks of overfitting and providing greater stability compared to E2E finetuning. Compared to task-agnostic sparse finetuning methods (LoRA and its variants~\cite{zhang2023adalora, valipour2022dylora}, heurstic parameter selection~\cite{neft}, or prefix-tuning~\cite{li2021prefix}), our approach enables focused adaptation to the classification task without adding any new parameters to the pretrained model. Past work supports our insight - domain-specific subwords reduce the need to pretrain from scratch on new domains~\cite{aws, zouhar2023tokenization, multigrain}.

Thus, in summary, we identify the most sensitive subset of model parameters that can be finetuned to seamlessly integrate task-specific supervised classification labels with the pretraining weights of the model. This includes representations for the newly introduced task-specific token constructs in \cref{fig:intro1}, as well as the most sensitive layer parameters and weights to adapt to the task. Our work thus addresses the limitations of both E2E finetuning and task-agnostic sparse finetuning approaches with a two-pronged solution: sparse finetuning synergized with task-specific tokenization for efficient adaptation of pretrained models. We make the following contributions in this work:

\begin{itemize}
\item We develop a novel sparse finetuning approach for specialized semantic classification tasks by leveraging task-specific token constructs (individual tokens, token sequences - both contiguous and non-contiguous) for parameter selection. In contrast to previous parameter-efficient finetuning approaches which do not customize per-task, we introduce task-specific tokenization and parameter selection to enable customized finetuning per classification task.

\item We extensively evaluate our proposed task-adaptation approach against three broad model adaptation approaches: end-to-end finetuning, low-rank adaptation, and prefix tuning. We consistently outperform these approaches on five diverse semantic classification tasks.
\end{itemize}

\section{Task-Specific Token Constructs}
\label{sec:vocab}

A token represents an atomic unit of semantic content in a language model \cite{webster1992tokenization}. A pretrained model requires two types of token updates for task adaptation. First, the semantics of the existing tokens in the model vocabulary need to be updated. For example, \textit{"therapy"} in generic use refers to mental health, while \textit{"gene therapy"} in medicine has a domain-specific semantic. Second, numerous domain or task-specific sequences of subwords are best represented as independent tokens. For the example in \cref{fig:intro1}, the terms \textit{BRCA} (gene family) and \textit{predisposition} are best represented as independent tokens rather than being partitioned into unrelated subwords. Further, we also note that some subsequences (both contiguous and non-contiguous) are ideally represented by separate tokens. For example, the sequences \textit{BRCA1 <-> cancer predisposition} and \textit{BRCA2 <-> DNA repair} in \cref{fig:intro1} represent gene-to-effect pairs. Representing each gene-to-effect pair as an independent token could help classify or identify the respective sentences appropriately. Thus, we introduce the above token constructs into the model vocabulary for task-adaptation. We now provide a theoretical analysis of tokenization for the finetuning objective in Section 2.1. We then develop a scalable approach to identify task-specific token constructs and subsequences in Section 2.2.

\subsection{Theoretical Interpretation}
\label{subsec:theory}
Finetuning a model for a given classification task optimizes the maximum likelihood optimization objective~\cite{yogatama2015bayesian} over labeled data points ($\mathbf{X}_i$, $y_{i}$) $\in$ $\mathbb{D}$ ($i = 1\ldots n$) where $\mathbb{D}$ denotes the classification task. During finetuning, the classifier model (with parameters $\theta$) receives the input tokens generated by the model tokenizer as \textit{tokenizer}($\mathbf{X}_i$) = $[x_{i}^{1}, x_{i}^{2} \ldots x_{i}^{k}]$ and the corresponding class label $y_i$ to optimize the following objective,
\begin{equation}
\mathbf{L_{MLE}}(\mathbf{\theta}) =  \sum_{i=1}^{n} \log P_{\theta}(y_i | x_{i}^{1}\ldots x_{i}^{k})
\label{mle}
\end{equation}

We now consider a perfect task-specific tokenizer (tokenizer{$_\mathbb{D}$}). Then, given the set of all possible labeled datapoints $(\mathbf{X}_i , y_i) \in \mathbb{D}$ and their true likelihood of appearing in the classification task $\mathbb{D}$, $P_{\mathbb{D}}(\mathbf{X}_i, y_i)$, and tokenizer{$_\mathbb{D}$}($\mathbf{X}_i$) = $[\chi_{i}^{1}, \ldots \chi_{i}^{k'}]$, the posterior for the classification task is given by,
\begin{equation}
\mathbf{L_{MAP}}(\mathbf{\theta}) = \hspace{-10pt} \sum_{(\mathbf{X}_i, y_i) \in \mathbb{D}} \hspace{-8pt} \log P_{\theta}(y_i | \chi_{i}^{1} \ldots \chi_{i}^{k'}) P_{\mathbb{D}}(\mathbf{X}_i, y_i)
\label{map}
\end{equation}

Clearly, a larger gap between the two tokenizations ($[x_{i}^{1} \ldots x_{i}^{k}]$ vs. $[\chi_{i}^{1}, \ldots \chi_{i}^{k'}]$) causes the posterior in \cref{map} and the MLE estimate in \cref{mle} to diverge resulting in a higher risk of overfitting and poor generalizability, especially with a smaller finetuning dataset. The larger the gap between the tokenizer vocabulary in \cref{mle} and the token distribution of the task, the greater the risk of overfitting. Thus, we aim to move the two distributions closer by adding task-specific token sequences to the tokenizer vocabulary.

\setlength{\textfloatsep}{5pt}

\begin{algorithm}[t]
\caption{Task-Specific Sequence Selection Algorithm (BIDE Closed-Sequence Mining Algorithm denoted $\text{BIDE-alg}$~\cite{bide})}
\begin{algorithmic}[1]
\State \textbf{Input:} Task-Specific Finetuning Dataset $\mathbb{D}$, Original Model Tokenizer $\textit{model-tok}$

\State $\text{Tokenized}_\mathbb{D} \gets \text{empty list}$
\ForAll{$d \in \mathbb{D}$}
        \State $\text{Tokenized}_\mathbb{D}$.add(\textit{model-tok}($d$))

    \EndFor

\noindent\rule{\linewidth}{0.2pt}
\Procedure{SequenceGen \text{  }} {$\text{min}\_\text{frequency}$, $\text{min}\_\text{len}$, $\text{max}\_\text{len}$}
    \State TaskSeqs $\gets$ $\text{empty dictionary}$

    \State candidates $\gets \text{BIDE-alg}$ ($\text{Tokenized}_\mathbb{D}$)
    
    \Comment{{\footnotesize\textit{filter by min\_frequency, min\_len, max\_len}}}
    
    \ForAll{seq $\in$ candidates}
        \State $\text{pplx}_{\text{seq}}$ $\gets$ $\textsc{ComputePPLX}(\text{seq})$
        
        \Comment{{\footnotesize\textit{seq is a sequence of tokens ($[x_{1} \ldots x_{k}]$)}}}
        
        
        \State {$\mathbf{if} \text{ } \text{pplx}_{\text{seq}}$  $\text{ }\& \text{ }$ $\text{freq}_{\text{seq}} \text{ }\geq \text{ } \text{cutoffs}$}

        \Comment{{\footnotesize\textit{cutoffs = [pplx-cutoff, freq-cutoff]}}}
        
        \State {\hspace{0\algorithmicindent} TaskSeqs[seq] $\gets$ new-id()}
        
        \Comment{{\footnotesize\textit{create a new tokenizer entry for seq}}}
        
    \EndFor
    \State \textbf{return} $\text{TaskSeqs}$
\EndProcedure

\vspace{4pt}
\Function{ComputePPLX}{$[x_{1} \ldots x_{k}]$}
    \State $p \gets ct(x_{1}) \text{ }/\text{ } {|\text{Tokenized}_\mathbb{D}|}$
    
    \Comment{\footnotesize{\textit{$ct(x)$ = frequency of x in $\text{Tokenized}_\mathbb{D}$}}\normalsize}
    
    \For{$j \gets 1$ to $k-1$}
    \vspace{3pt}
        \State $P_{\mathbb{D}}(x_{j+1} | x_{j})$ = $ct(x_{j} x_{j+1}) \text{ } / \text{ } ct(x_{j})$ 
        
   \Comment{\footnotesize{\textit{$ct(xy)$ = frequency of bigram xy in $\text{Tokenized}_\mathbb{D}$}}\normalsize}
   
    \vspace{2pt}
        \State $p \gets p \times$ $P_{\mathbb{D}}(x_{j+1} | x_{j})$
    \EndFor
    \State \Return $p^{-\frac{1}{k}}$
\EndFunction


            
            

\label{alg:phrase}
\end{algorithmic}
\end{algorithm}

\subsection{Task-Specific Token Selection}
\label{subsec:tokens}

To align the model tokenizer with the ideal task-specific tokenization introduced in \cref{subsec:theory}, we first identify task-specific contiguous and non-contiguous sequences of subwords in the finetuning dataset $\mathbb{D}$. This helps us better represent the task-specific token distribution in \cref{map}. 

First, we generate the original model tokenizer outputs for every input in the finetuning dataset $\mathbb{D}$ as described in \cref{alg:phrase}. We then mine frequent token sequences from these token sequences via bidirectional closed sequence mining \cite{bide, han2007frequent}. A bidirectional closed sequence does not have any subsequence or super-sequence with the same frequency~\cite{bide}, and includes both contiguous and non-contiguous sequences of tokens. For example, \textit{"gene therapy"} is a closed sequence since super-sequences such as \textit{"in-situ gene therapy"} have lower frequencies, while subsequences like \textit{"gene"} have higher frequencies. We remove candidate sequences that are too long ($\geq$ 20 tokens) and evaluate the rest with two metrics, sequence frequency, and the unigram-normalized perplexity~\cite{roh2020unigram}. We describe the details of the metrics and the sequence selection algorithm in \cref{alg:phrase}. We choose the frequency and perplexity cut-off thresholds so that the number of newly added task-specific sequences is no more than 10\% of the model vocabulary, i.e., at most 10\% of all tokens are task-specific.

After the sequence selection is complete, we assign new token ids to the selected sequences and randomly initialize the corresponding token embeddings in the model. The modified tokenization works as follows - given an input text, as described in \cref{alg:tok} (\textsc{TaskTokenizer}), we first tokenize the input with the original model tokenizer and identify the task-specific candidate sequences that may be present in this input using an index lookup on each token. For example, given the token \textit{cancer} for an input text, the candidates for the task-specific sequence are \textit{cancer predisposition} and \textit{cancer treatment}. If either of these sequences is present in the input, we insert the corresponding new token id assigned to the sequence at its start position(s) within the token list. The model receives these combined tokens (including the original tokens and sequence tokens) for both finetuning (described in Section 3) and at inference time.

\begin{algorithm}[t]
\caption{Task-Adaptive Tokenization}
\begin{algorithmic}[1]
\State \textbf{Input:} $\text{TaskSeqs}$ from \textsc{SequenceGen} in Algorithm 1, Original model tokenizer $\textit{model-tok}$

\State Index $\gets$ $\text{empty dictionary}$
\State NewIds $\gets$ $\text{empty dictionary}$
\ForAll{$\text{seq} \in \text{TaskSeqs}$}
    \State NewIds[seq] $=$ new-token-id()

    \Comment{{\footnotesize\textit{Create a new token id in the model vocabulary -}}}
    
    \Comment{{\footnotesize\textit{and initialize an embedding for each sequence}}}
    
    \ForAll{$\text{token} \in \text{seq}$}
        \State Index[token].\textit{add}((seq, NewIds[seq]))
        
      \EndFor
\EndFor 

\noindent\rule{\linewidth}{0.2pt}

\Statex \textbf{Procedure:} \textsc{TaskTokenizer}\hspace{2pt}(input-seq)

\State tokens $ \gets $ \textit{model-tok}\hspace{1pt}(input-seq)


\ForAll{$\text{t} \in \text{tokens}$}
    \ForAll {$\text{seq, seq-id} \in \text{Index[t]}$}
    \If{seq \textbf{in} tokens}
                \State tokens.\textit{insert} (seq-id)
    
                \Comment{{\footnotesize\textit{append the seq-id token for seq at the}}}
                
                \Comment{{\footnotesize\textit{start-position(s) of seq} \textit{within tokens}}}
                
            \EndIf
    \EndFor
\EndFor
\vspace{2pt}
\label{alg:tok}
\end{algorithmic}
\vspace{4pt}
\end{algorithm}

\section{Task-Adaptive Sparse Finetuning}
\label{sec:sparse}
In the previous section, we selected task-specific token sequences and added them to the vocabulary as new tokens. We randomly initialize the token embeddings for these new tokens. In this section, we develop our sparse finetuning approach to select sensitive model parameters which we finetune along with the new tokens to adapt the model to the classification task.

We measure parameter sensitivity on the finetuning data $\mathbb{D}$ with a direct metric and a gradient-based metric. The direct metric measures the variance observed for the respective outputs across the finetuning data. This enables us to select the parameters that are most receptive to diverse inputs in the dataset. Specifically, given a set of inputs for the classification task ($\mathbf{X}_i$, $y_{i}$) $\in$ $\mathbb{D}$, we measure the index of dispersion for each module across these inputs. A module $\mathcal{M}$ refers to a functional unit of computation such as a feed-forward layer or an attention block. In this case, we measure the output of each module, $\mathcal{M}$, as follows:

\begin{equation}
    \mathbb{I}_{\mathcal{M}}(\mathbb{D}) = \frac{\sigma^2(\mathcal{M}(\mathbb{D}))}{\mu(\mathcal{M}(\mathbb{D}))}
\end{equation}

where $\sigma^2(\mathcal{M}(\mathbb{D}))$ is the variance of the module outputs and $\mu(\mathcal{M}(\mathbb{D}))$ is the mean of the module outputs over all inputs $\mathbf{X}_i$ $\in$ $\mathbb{D}$. This index of dispersion helps us identify the parameters that show significant variation across the dataset $\mathbb{D}$ indicating their sensitivity and importance for finetuning. We also measure the gradient norm for each module,

\begin{equation}
  \mathbb{G}_{\mathcal{M}}(\mathbf{X}_i, y_i) =  \|\nabla \mathcal{L}(\mathcal{M}(\mathbf{X}_i), y_i)\|
\end{equation}

where $\nabla \mathcal{L}$ denotes the gradient of the loss function $\mathcal{L}$ with respect to the module parameters for the datapoint $(\mathbf{X}_i, y_i)$. We then average the gradient norm across all finetuning datapoints:

\begin{equation}
    \mathbb{G}_{\mathcal{M}}(\mathbb{D}) = \frac{1}{|\mathbb{D}|} \sum_{(\mathbf{X}_i, y_i) \in \mathbb{D}} \mathbb{G}_{\mathcal{M}}(\mathbf{X}_i, y_i)
\end{equation}

By combining both the index of dispersion and the average gradient norm over the entire dataset, we robustly identify the most sensitive parameters for finetuning. We used a linear combination of these two metrics to compute the sensitivity of a specific module $\mathcal{M}$ over dataset $\mathbb{D}$ as follows:

\begin{equation}
    \mathbb{S}_\mathcal{M}(\mathbb{D}) = \mathbb{I}_{\mathcal{M}}(\mathbb{D}) + \beta \cdot \mathbb{G}_{\mathcal{M}}(\mathbb{D})
    \label{eq:sensitivity}
\end{equation}

where $\beta$ controls the relative importance of the index of dispersion and the gradient norm. We tune coefficient $\beta$ per dataset via grid search (details in \cref{appendix:res}), and select modules with the highest values of $\mathbb{S}_\mathcal{M}(\mathbb{D})$ for finetuning over $\mathbb{D}$.

By computing the sensitivity $\mathbb{S}_\mathcal{M}(\mathbb{D})$ in \cref{eq:sensitivity} for each module, we are able to prioritize parameters that are responsive to different inputs (measured by $\mathbb{I}_{\mathcal{M}}(\mathbb{D})$) and also prioritized by the optimizer when finetuning (measured by $\mathbb{G}_{\mathcal{M}}(\mathbb{D})$). This combination enables balanced parameter selection for a given dataset or task.

\section{Experiments}
\label{sec:ne_exp_settings}

\begin{table*}[ht]
  \caption{Comparison of our approach (AdaPT) against E2E finetuning (FT), AdaLoRA (AL), prefix tuning (PT), and Layer Selection (LS) across all datasets. For our approach, we choose to tune 10\% of the parameters for the four larger models (BERT large/base, Gemma, Phi-3.5) and 20\% for the two smaller models (BERT medium/small). We note that our approach shows stronger gains for the more specialized tasks (RCT, ACL-ARC, SciERC) vs. the rest (hyperpartisan, movie reviews). We bold the highest performance in each dataset/model combination and underline if two values are close.}
  \label{tab:bert-e2e}
  \centering
  \small
  \setlength{\tabcolsep}{4.6pt} 
  \vspace{-3pt}
  \begin{tabular}{c|ccccc|ccccc|ccccc}
\toprule
Task & \multicolumn{5}{c}{BERT-large} & \multicolumn{5}{c}{BERT-base} & \multicolumn{5}{c}{BERT-medium} \\
\cmidrule(lr){2-6} \cmidrule(lr){7-11} \cmidrule(lr){12-16}
& FT & AdaPT & AL & PT & LS & FT & AdaPT & AL & PT & LS & FT & AdaPT & AL & PT & LS \\
\midrule
Hyperpartisan & 0.892 &\textbf{0.908}& 0.828 & 0.803 & 0.816 & 0.861 &\textbf{0.877}& 0.794 & 0.759 & 0.813 & 0.840 &\textbf{0.872}& 0.776 & 0.725 & 0.761 \\
RCT           & 0.805 &\textbf{0.824}& 0.787 & 0.716 & 0.799 & 0.793 &\textbf{0.812}& 0.770 & 0.692 & 0.755 & 0.784 &\textbf{0.811}& 0.761 & 0.680 & 0.752 \\
Movie Reviews & \underline{0.934} &\underline{0.937}& 0.922 & 0.929 & 0.907 & \underline{0.918} &\underline{0.916}& 0.904 & 0.890 & 0.892 & \underline{0.911} &\underline{0.907}& 0.877 & 0.886 & 0.883 \\
ACL-ARC       & 0.723 &\textbf{0.739}& 0.689 & 0.648 & 0.682 & 0.720 &\textbf{0.732}& 0.667 & 0.614 & 0.675 & 0.695 &\textbf{0.716}& 0.623 & 0.597 & 0.608 \\
SciERC        & 0.791 &\textbf{0.810}& 0.722 & 0.705 & 0.710 & 0.769 &\textbf{0.785}& 0.698 & 0.629 & 0.701 & 0.713 &\textbf{0.741}& 0.693 & 0.634 & 0.687 \\
\midrule
Task & \multicolumn{5}{c}{BERT-small} & \multicolumn{5}{c}{Google Gemma} & \multicolumn{5}{c}{Microsoft Phi-3.5} \\
\cmidrule(lr){2-6} \cmidrule(lr){7-11} \cmidrule(lr){12-16}
& FT & AdaPT & AL & PT & LS & FT & AdaPT & AL & PT & LS & FT & AdaPT & AL & PT & LS \\
\midrule
Hyperpartisan & 0.816 &\textbf{0.835}& 0.772 & 0.751 & 0.789 & 0.885 & \textbf{0.901} & 0.812 & 0.785 & 0.797 & 0.855 & \textbf{0.891} & 0.798 & 0.744 & 0.782 \\
RCT           & 0.772 &\textbf{0.788}& 0.743 & 0.710 & 0.761 & 0.809 & \textbf{0.827} & 0.773 & 0.738 & 0.757 & 0.799 & \textbf{0.819} & 0.755 & 0.682 & 0.739 \\
Movie Reviews & \underline{0.892} &\underline{0.901}& 0.868 & 0.881 & 0.850 & \underline{0.925} & \underline{0.931} & 0.912 & 0.903 & 0.898 & \underline{0.919} & \underline{0.923} & 0.891 & 0.904 & 0.901 \\
ACL-ARC       & 0.682 &\textbf{0.711}& 0.586 & 0.554 & 0.599 & 0.727 & \textbf{0.741} & 0.681 & 0.658 & 0.667 & 0.711 & \textbf{0.728} & 0.649 & 0.595 & 0.638 \\
SciERC        & 0.706 &\textbf{0.723}& 0.642 & 0.603 & 0.617 & 0.785 & \textbf{0.802} & 0.711 & 0.606 & 0.726 & 0.770 & \textbf{0.795} & 0.681 & 0.613 & 0.676 \\
\bottomrule
\end{tabular}
  \vspace{-4pt}
\end{table*}

\begin{table*}[ht]
  \caption{We further test our approach by finetuning 5, 10, 15\% or 20\% of the model parameters in each dataset, outperforming E2E finetuning and AdaLoRA in \cref{tab:bert-e2e}. We find that finetuning ~10\% parameters yields close to optimal performance for larger models, while 20\% is better for smaller models. We do not observe noticeable gains when tuning more than 20\% parameters. We tablulate the results for 50\% and 100\% (full) parameter tuning in \cref{tab:tab-bert-appendix}.}
  \label{tab:bert-adapt}
  \centering
  \small 
  \begin{tabular}{c|cccc|cccc|cccc}
    \toprule
    Task & \multicolumn{4}{c}{BERT-large AdaPT} & \multicolumn{4}{c}{BERT-base AdaPT} & \multicolumn{4}{c}{BERT-medium AdaPT} \\
    \cmidrule(lr){2-5} \cmidrule(lr){6-9} \cmidrule(lr){10-13}
          & 20\% & 15\% & 10\% & 5\% & 20\% & 15\% & 10\% & 5\% & 20\% & 15\% & 10\% & 5\% \\
    \midrule
    Hyperpartisan & \underline{0.913} & \underline{0.914} & 0.908 & 0.825 & \underline{0.874} & 0.871 & \underline{0.877} & 0.796 & \textbf{0.872} & 0.862 & 0.865 & 0.813 \\
    RCT           & \underline{0.827} & 0.822 & \underline{0.824} & 0.761  & 0.805 & 0.799 & \textbf{0.812} & 0.761 & \textbf{0.811} & 0.805 & 0.798 & 0.746 \\
    Movie Reviews & \underline{0.939} & \underline{0.943} & 0.937 & 0.893 & \textbf{0.924} & 0.914 & 0.916 & 0.872 & \underline{0.907} & \underline{0.904} & 0.879 & 0.883 \\
    ACL-ARC       & 0.730 & 0.728 & \textbf{0.739} & 0.685 & \underline{0.729} & 0.717 & \underline{0.732} & 0.691 & \underline{0.716} & \underline{0.718} & 0.699 & 0.683 \\
    SciERC        & \underline{0.808} & 0.803 & \underline{0.810} & 0.744 & \underline{0.790} & \underline{0.792} & 0.785 & 0.738 & \underline{0.741} & \underline{0.739} & 0.723 & 0.645 \\
    \midrule
    Task & \multicolumn{4}{c}{BERT-small AdaPT} & \multicolumn{4}{c}{Google Gemma AdaPT} & \multicolumn{4}{c}{Microsoft Phi-3.5 AdaPT} \\
    \cmidrule(lr){2-5} \cmidrule(lr){6-9} \cmidrule(lr){10-13}
          & 20\% & 15\% & 10\% & 5\% & 20\% & 15\% & 10\% & 5\% & 20\% & 15\% & 10\% & 5\% \\
    \midrule
    Hyperpartisan & \textbf{0.835} & 0.824 & 0.821 & 0.795 & \underline{0.904} & 0.892 & \underline{0.901} & 0.878 & 0.885 & \textbf{0.896} & 0.891 & 0.809 \\
    RCT           & \underline{0.788} & \underline{0.790} & 0.775 & 0.735 & 0.818 & \underline{0.831} & \underline{0.827} & 0.795 & \textbf{0.822} & 0.812 & 0.819 & 0.780 \\
    Movie Reviews & \textbf{0.901} & 0.886 & 0.889 & 0.867 & \underline{0.932} & \underline{0.935} & 0.931 & 0.889 & \underline{0.922} & 0.917 & \underline{0.923} & 0.881 \\
    ACL-ARC       & \textbf{0.711} & 0.695 & 0.672 & 0.678 & 0.734 & 0.737 & \textbf{0.741} & 0.722 & 0.724 & \underline{0.727} & \underline{0.728} & 0.709 \\
    SciERC        & \textbf{0.723} & 0.712 & 0.694 & 0.632 & \textbf{0.807} & 0.798 & 0.802 & 0.765 & \underline{0.793} & 0.778 & \underline{0.795} & 0.745 \\
    \bottomrule
  \end{tabular}
  \vspace{-8pt}
\end{table*}

\begin{table*}[ht]
  \caption{We do not observe a noticeable advantage when tuning more than 20\% of the BERT model parameters for the chosen tasks. In some tasks, tuning more parameters leads to a drop indicating potential overfitting.}
  \label{tab:tab-bert-appendix}
  \centering
  \small 
  \begin{tabular}{c|ccc|ccc|ccc|ccc}
    \toprule
    Task & \multicolumn{3}{c}{BERT-large AdaPT} & \multicolumn{3}{c}{BERT-base AdaPT} & \multicolumn{3}{c}{BERT-medium AdaPT} & \multicolumn{3}{c}{BERT-small AdaPT} \\
    \cmidrule(lr){2-4} \cmidrule(lr){5-7} \cmidrule(lr){8-10} \cmidrule(lr){11-13}
          & 20\% & 50\% & 100\% & 20\% & 50\% & 100\% & 20\% & 50\% & 100\% & 20\% & 50\% & 100\% \\
    \midrule
    Hyperpartisan & \textbf{0.913} & 0.894 & 0.901 & \underline{0.874} & \underline{0.878} & 0.869 & \underline{0.872} & \underline{0.868} & \underline{0.874} & \underline{0.835} & \underline{0.828} & 0.817 \\
RCT           & \underline{0.827} & \underline{0.832} & 0.819  & \underline{0.799} & 0.783 & \underline{0.796} & \underline{0.811} & 0.797 & \underline{0.804} & \underline{0.788} & \underline{0.790} & 0.782 \\
Movie Reviews & \textbf{0.939} & 0.933 & 0.928 & \underline{0.924} & \underline{0.926} & 0.919 & \underline{0.907} & \underline{0.908} & 0.898 & \underline{0.901} & 0.890 & 0.895 \\
ACL-ARC       & \underline{0.730} & \underline{0.731} & \underline{0.730} & \underline{0.726} & \underline{0.724} & 0.718 & \textbf{0.716} & 0.702 & 0.694 & \underline{0.711} & 0.698 & \underline{0.705} \\
SciERC        & \textbf{0.808} & 0.801 & 0.794 & \textbf{0.790} & 0.787 & 0.778 & \textbf{0.741} & 0.731 & 0.729 & \underline{0.723} & 0.718 & \underline{0.727} \\
    \bottomrule
  \end{tabular}
\end{table*}

We tested our approach on small language models (SLMs) Microsoft Phi-3.5 Instruct~\cite{phi3}, Google Gemma-2 2B Instruct~\cite{gemma} and \textit{BERT} pretrained encoders\footnote{https://github.com/google-research/bert} (\textit{BERT-large, BERT-base, BERT-medium, and BERT-small}) on five specialized semantic classification tasks: Politics (HyperPartisan\footnote{https://paperswithcode.com/dataset/hyperpartisan}), biomedical (RCT\footnote{https://paperswithcode.com/dataset/pubmed-rct}), computer science (ACL-ARC\footnote{https://paperswithcode.com/sota/citation-intent-classification-on-acl-arc}), scientific content (SciERC\footnote{https://paperswithcode.com/sota/joint-entity-and-relation-extraction-on}), and movies (RT Movie Reviews\footnote{https://www.kaggle.com/datasets/stefanoleone992/rotten-tomatoes-movies-and-critic-reviews-dataset}). We compare the following methods:

\noindent \textbf{Our Approach (AdaPT)} - \underline{Ad}aptive Tokenization \underline{a}nd Sparse \underline{P}arameter \underline{T}uning) - In each dataset, we expand the tokenizers of each model with our selected sequences, and finetune with the new tokens as described in \cref{sec:sparse}. We provide a case study of sample sequences in \cref{appendix:tokens}. We vary the count of model parameters selected for finetuning to 5\%, 10\%, 15\% and 20\% of all parameters for each model.

\noindent \textbf{End-to-End (E2E) Finetuning}~\cite{sun2019fine} - We finetune all parameters of each model with the finetuning datasets.

\noindent \textbf{AdaLoRA}~\cite{hu2021lora, zhang2023adalora} - We test AdaLoRA on each dataset and set the rank budget so that the number of parameters tuned is close to 20\% (largest setting for our approach). 


\noindent \textbf{Layer Selection}~\cite{pan2024lisa} - We test layer selection on each dataset and set the layer count so that the number of parameters tuned is close to 20\% (largest setting for our approach). 

\noindent \textbf{Prefix Tuning}~\cite{li2021prefix} - We test prefix tuning on each dataset and set the number of prefix tokens to 20\% of the average sequence length.

\noindent \textbf{0-shot / 10-shot Prompting} - We evaluate zero and 10-shot performance of Microsoft Phi-3.5-Instruct and Google Gemma-Instruct, compared to the OpenAI GPT-4 model. This provides us a baseline for finetuning-based methods. We share our prompts for each of the five tasks in \cref{appendix:llm}. 

\subsubsection{Metrics}

We measure the accuracy and stability of each method on each task with the following two metrics. 

\begin{figure*}[t]
  \centering
    
\includegraphics[width=0.86\textwidth]{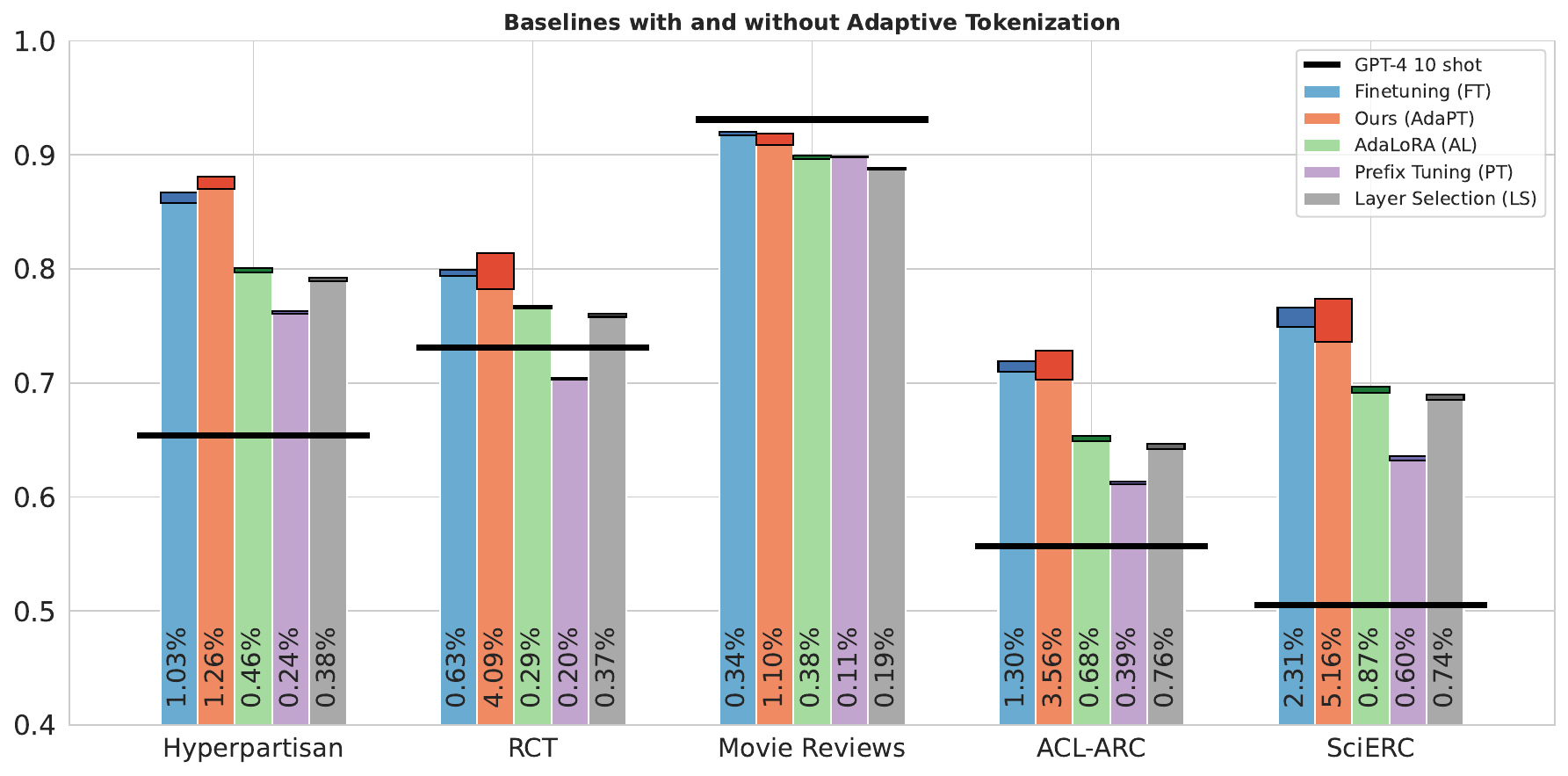}
\vspace{-11pt}
\caption{We tested each method (including ours) with and without task-specific tokenization on all six models and measure the average accuracy result. At the bottom of each bar, we specify the relative percentage improvement with task-specific tokenization. The black lines indicate 10-shot OpenAI GPT-4 accuracy on the task. Task-specific tokenization was most effective with our method, but also improves E2E finetuning. However, the impact of task tokens on prefix tuning, layer selection, and AdaLoRA are not significant.}
  \label{fig:ablation}
  
\end{figure*}
\noindent \textbf{Accuracy Metric: } We measure the F1-score for each classification task. For datasets with more than two classes, we compute the averaged F1-score over all classes (\cref{tab:bert-adapt}, \cref{tab:bert-e2e}).

\noindent \textbf{Stability Metric: } We evaluate the variance of the average F1-score for the non-LLM baselines over all datasets with shuffled data, different random initializations, and train-test variations in \cref{tab:bert-stability}.

\begin{table}[b]
  
  \label{tab:llm}
  \small
  \setlength{\tabcolsep}{3.6pt}
  \centering
  \begin{tabular}{c|cc|cc|cc}
    \toprule
    Task  & \multicolumn{2}{c}{OpenAI GPT-4} & \multicolumn{2}{c}{Gemma-2 2B IT} & \multicolumn{2}{c}{Msft Phi-3.5 IT} \\
    \cmidrule(lr){2-3} \cmidrule(lr){4-5} \cmidrule(lr){6-7}
           & 0 & 10 & 0 & 10  & 0 & 10 \\
    \midrule
    Hyper & 0.544 & \textbf{0.654} & 0.430 & 0.508 & 0.492 & 0.595 \\
    RCT           & 0.636 & \textbf{0.731} & 0.384 & 0.624 & 0.482 & 0.576 \\
    Movie          & \textbf{0.974} & 0.931 & 0.863 & 0.898 & 0.902 & 0.925 \\
    ACL       & 0.525 & \textbf{0.557} & 0.479 & 0.508 & 0.505 & 0.540 \\
    Sci        & 0.443 & \textbf{0.505} & 0.314 & 0.396 & 0.387 & 0.459 \\
    \bottomrule
  \end{tabular}
  \vspace{2pt}
  \caption{Zero and 10-shot Prompting for the Gemma and Phi-3.5 SLMs, and OpenAI GPT-4. We share the prompts in \cref{appendix:llm}.}
  \vspace{-18pt}
\end{table}



\subsection{Results}

We compare our approach with the baseline finetuning methods in \cref{tab:bert-e2e}, and further benchmark our approach when varying the percentage of finetuned parameters in \cref{tab:bert-adapt}. Our approach proves effective in specialized academic or scientific tasks where finetuning is crucial (SciERC, ACL-ARC, RCT). AdaLoRA and prefix tuning are effective for the two tasks where the pretraining is more likely to cover the task (movie reviews, hyperpartisan). Layer-Selection was comparable to AdaLoRA on most datasets. End-to-end finetuning was effective across all tasks but underperforms our approach combining adaptive tokenization with sparse finetuning, especially on the more specialized tasks. We share qualitative details on the tokens selected by our approach in \cref{appendix:tokens}.

For further clarity, we also combined each baseline method with our adaptive tokenization approach and provide a detailed ablation analysis in \cref{subsec:ablation2}. Our approach stood out for its adaptability, offering a viable option for all scenarios. We note that our performance gains saturate when tuning 10-20\% of the model parameters. We present additional results for 50\% and 100\% parameter tuning in \cref{appendix:res}. We also evaluate the two SLM models (Gemma, Phi2) and one SoTA LLM (OpenAI GPT-4) on all five datasets with zero-shot and ten-shot prompts in \cref{tab:llm} to provide a baseline for all finetuning methods. We share the prompts in \cref{appendix:llm}. We find zero/few-shot prompting to underperform other baselines on every dataset except movie reviews indicating the importance of model adaptation for specialized classification tasks.

\vspace{-5pt}
\subsection{Ablation Analysis}
\label{subsec:ablation2}
In this section, we present a detailed empirical ablation analysis of each finetuning baseline with and without introducing task-specific tokens in \cref{fig:ablation}. This helps us to understand the contribution of each aspect to the final result. For our approach, AdaPT, we attempt a variation where we apply the sparse finetuning technique without the new task-specific tokens. For end-to-end finetuning and the other baselines, we introduce the same task-specific tokens selected by our approach at the finetuning stage. For AdaLoRA, prefix tuning, and layer selection, we also allow gradient updates for the token embeddings of the new tokens.

We document each variation with and without the new task-specific tokens in \cref{fig:ablation}. At the bottom of each bar, we indiciate the relative percentage improvement observed from introducing the task-specific tokens. All results are averaged over all six models (BERT-variations, Google Gemma, Microsoft Phi 2). AdaLoRA, prefix-tuning and layer selection show limited benefit from the new tokens due to their lack of adaptive parameter selection. End-to-end finetuning benefits from the new tokens with smaller relative gains compared to our approach. One likely explanation is the tendency to overfit with end-to-end finetuning, which could explain the reduced gains in some tasks. We do not observe a noticeable difference in the movie-reviews task. 

\subsection{Computation Costs}
\label{subsec:cost}

\begin{figure}[t]
\caption{We compare the finetuning and inference times for BERT-large with each method for one epoch with 5000 samples, batch size 5 on a single Nvidia A100 GPU, averaged across all five datasets. Each bar displays the percentage change compared to finetuning. Our approach is competitive with AdaLoRA and Prefix Tuning for training and faster at inference time since it does not add any parameters.}
  \centering
\includegraphics[width=0.47\textwidth]{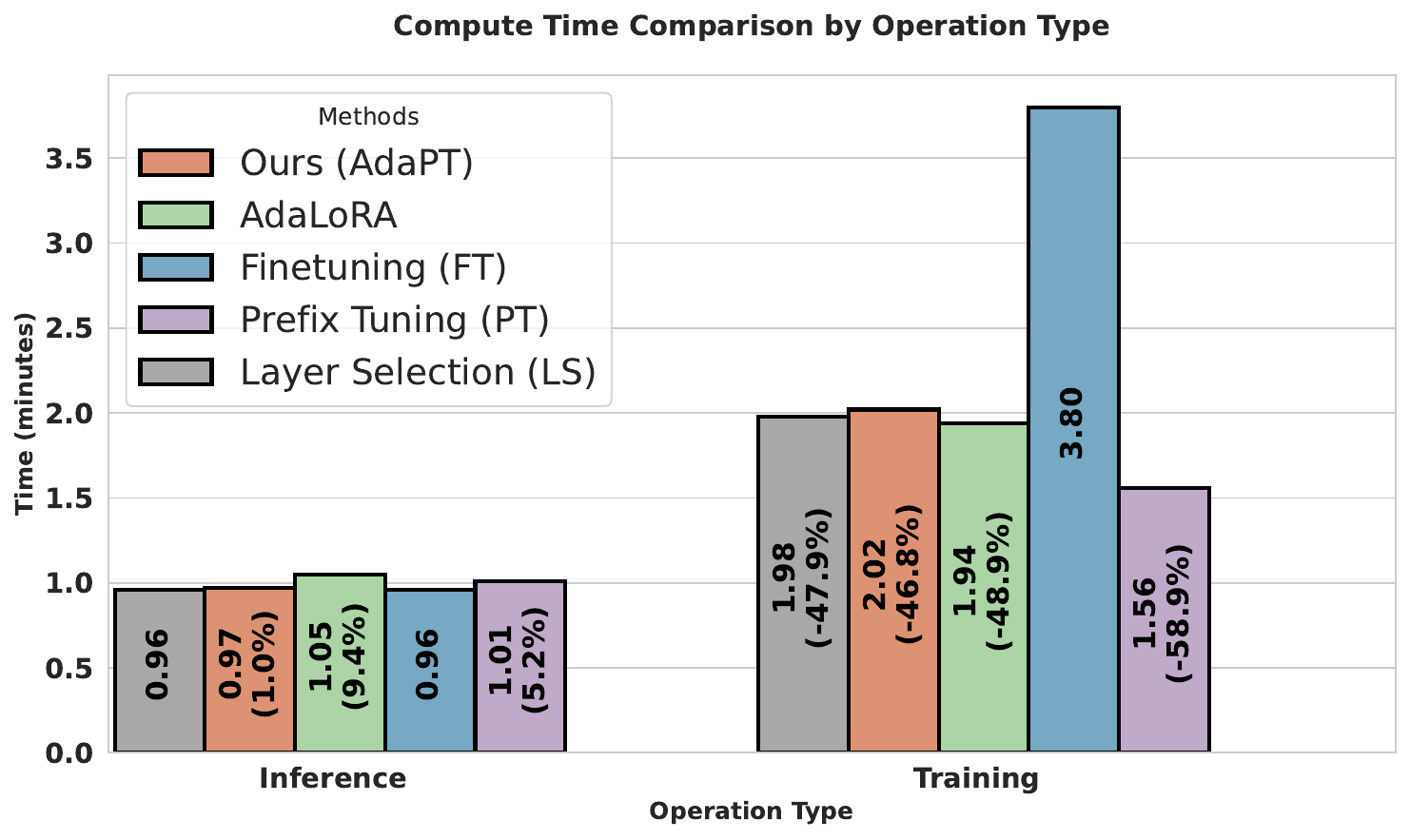}
\vspace{-2pt}
  \label{fig:compute}

\end{figure}

We compare the per-epoch training costs for our method, as well as the three baselines - End-to-end finetuning, AdaLoRA, and Prefix-Tuning. We apply each method to the BERT-large checkpoint (330 M parameters) on five-thousand samples from each of the five datasets and plot the averages in \cref{fig:compute}.  For training, We find that our method is roughly half the cost of end-to-end finetuning despite the addition of new tokens, and comparable to AdaLoRA and Prefix Tuning. For inference, our approach is 5-8\% cheaper than AdaLoRA and Prefix-Tuning since it does not add any new parameters to the model at inference time. Our approach incurs a slight overhead (by about 1\%, 0.97 vs. 0.96 minutes) compared to the purely finetuned model at inference time due to the additional tokens.

\begin{table}[b]
  \label{tab:flops}
  \centering
  \small
  \begin{tabular}{cccccc}
    \toprule
    Task & FT & AdaPT (20\%) & AL & PT & LS\\
    \midrule
    TFlops per Epoch & 2486 & 1104 & 1314 & 982 & 1071 \\
    Peak GPU Mem (GB)  & 21.8 & 14.2 & 17.1 & 12.6 & 13.9 \\
    \bottomrule
  \end{tabular}
  \vspace{2pt}
  \caption{We measure the average flops (in trillions of flops) per training epoch and peak GPU memory usage (GB) for the BERT-large checkpoint on a single A100 GPU, averaged over the five datasets. Our method is significantly more efficienct than E2E finetuning, and competitive with the other baselines on both metrics.}
  \vspace{-15pt}
\end{table}

\subsection{Stability}

End-to-end finetuning is known to produce a realtively large variance depending on initialization and random factors such as data order. To evaluate stability, we performed 10 training runs of each method with different random initializations of the new parameters (new token embeddings for our method, LoRA parameters, and the prompt in prompt tuning). We report our observed variance in the F1 scores in \cref{tab:bert-stability}. Our method exhibits a noticeably smaller variance compared to end-to-end finetuning (by about 50\%).

\begin{table}[t]
  \caption{We measure the stability of each baseline on all five datasets. We measure the standard deviation of the F1-score across 20 training runs with different initializations for each method. We observe our method to show a lower variance compared to E2E finetuning. Notably AdaLoRA, LS and PT show lower variances than our method, which potentially indicates a lower degree of adaptation.}
  \label{tab:bert-stability}
  \centering
  \small
  \begin{tabular}{cccccc}
    \toprule
    Task & FT & AdaPT & AL & PT & LS \\
    \midrule
    Hyperpartisan & 0.045 & 0.026 & 0.012 & 0.009 & 0.017  \\
    RCT           & 0.218 & 0.167 & 0.102 & 0.125 & 0.119 \\
    Movie Reviews          & 0.024 & 0.009 & 0.004 & 0.010 & 0.008 \\
    ACL-ARC       & 0.064 & 0.041 & 0.025 & 0.031 & 0.030 \\
    SciERC        & 0.087 & 0.038 & 0.016 & 0.010 & 0.025 \\
    \bottomrule
  \end{tabular}
\end{table}



\section{Case-Study on the Selected Tokens}

\label{appendix:tokens}

We provide an analysis of how our approach adapts to each domain by grouping together sequences. Our method identifies coherent sequences in each dataset that should be combined into a single new token, while the BERT tokenizer splits these sequences into multiple tokens.

\subsection{Hyperpartisan Dataset}




Our model groups together numerous sequences that help correctly classify political content. In each row of the table, we specify the identified sequence, and the corresponding tokens of the BERT tokenizer. Our method merges these sequences into single new tokens. We share a few samples below.

\begin{center}
\small
\begin{tabular}{ll}
\hline
\textbf{Identified Patterns} & \textbf{BERT tokenizer output} \\
\hline
\multicolumn{2}{l}{\textit{Token Combining for Named Entities}} \\
\hline
New York City & [2047, 2259, 2103] \\
Barack Obama & [13857, 8112] \\
Donald Trump & [6221, 8398] \\
Vladimir Putin & [8748, 22072] \\
\hline

\multicolumn{2}{l}{\textit{Financial and Market Terminology}} \\
\hline
Wall Street Journal & [2813, 2395, 3485] \\
Dow Jones Industrial Average & [23268, 3557, 3919, 2779] \\
NASDAQ Composite & [17235, 2850, 4160, 12490] \\
Stock Advisor & [4518, 8619] \\
fourth quarter & [2959, 16211, 19418] \\
\hline
\multicolumn{2}{l}{\textit{Political and Governmental Terms}} \\
\hline
White House Press Secretary & [2317, 2160, 2811, 3187] \\
US President Donald Trump & [2149, 2343, 6221, 8398] \\
Senate Majority Leader & [4001, 3484, 3003] \\
Government shutdown & [2231, 3844, 7698] \\
Affordable Care Act & [15184, 2729, 2552] \\
\hline
\multicolumn{2}{l}{\textit{News Source Attributions}} \\
\hline
Wall Street Journal & [2813, 2395, 3485] \\
New York Times & [2047, 2259, 2335] \\
MarketWatch & [3006, 18866] \\
Motley Fool & [9587, 18492, 7966] \\
\hline
\end{tabular}
\end{center}

\subsection{ACL-ARC and Sci-ERC Datasets}
Our model groups together linguistic structures and numerical references that are split by the BERT tokenizer into single tokens.

\begin{center}
\small
\begin{tabular}{lcl}
\hline
\textbf{Identified Patterns} & \textbf{BERT tokenizer output} \\
\hline
\multicolumn{2}{l}{\textit{Numerical References in Academic Writing}} \\
\hline
Number (nu, \#\#m) & [16371, 2213] \\
Table (table, nu, \#\#m) & [2795, 16371, 2213] \\
Figure (figure, nu, \#\#m) & [3275, 16371, 2213] \\
Section (section, nu, \#\#m) & [2930, 16371, 2213] \\
\hline
\multicolumn{2}{l}{\textit{Technical Linguistic Terminology}} \\
\hline

syntactic & [19962, 2696, 13306] \\
lexical & [16105, 9289] \\
probabilistic & [4013, 3676, 27965, 4588] \\
antecedent & [14405, 26005, 16454] \\
annotated & [5754, 17287, 3064] \\
part of speech & [2112, 1997, 4613] \\
\hline

\multicolumn{2}{l}{\textit{Common Academic Phraseology}} \\
\hline

according to the & [2429, 2000, 1996] \\
in terms of & [1999, 3408, 1997] \\
with respect to & [2007, 4847, 2000] \\
in this paper we & [1999, 2023, 3259, 2057] \\
such as the & [2107, 2004, 1996] \\
can be used & [2064, 2022, 2109] \\
based on the & [2241, 2006, 1996] \\
there is no & [2045, 2003, 2053] \\
\hline

\multicolumn{2}{l}{\textit{Reference Markers \& Connective Phrases}} \\
\hline

e.g. & [1041, 1012, 1043, 1012] \\
i.e. & [1045, 1012, 1041, 1012] \\
on the other hand & [2006, 1996, 2060, 2192] \\
a number of & [1037, 2193, 1997] \\
the fact that & [1996, 2755, 2008] \\
in order to & [1999, 2344, 2000] \\
\hline
\end{tabular}
\end{center}

\subsection{Movie-Reviews Dataset}
Our model combines tokens in movie review texts for ratings, evaluative language, and cinema-specific terminology. We present a few sample phrases below.

\begin{center}
\small
\begin{tabular}{lcl}
\hline
\textbf{Identified Pattern} & \textbf{BERT tokenizer output} \\
\hline
\multicolumn{2}{l}{\textit{Numeric Ratings}} \\
\hline
7 out of 10 & [1021, 2041, 1997, 2184] \\
2/10 & [1016, 1013, 2184] \\
10/10 & [2184, 1013, 2184] \\
\hline
\multicolumn{2}{l}{\textit{Star Ratings}} \\
\hline
***** & [1008, 1008, 1008, 1008, 1008] \\
\hline
\multicolumn{2}{l}{\textit{Opening Phrases}} \\
\hline
in this movie & [1999, 2023, 3185] \\
one of the best & [2028, 1997, 1996, 2190] \\
this is a & [2023, 2003, 1037] \\
\hline
\multicolumn{2}{l}{\textit{Critical Assessments}} \\
\hline
could have been & [2071, 2031, 2042] \\
don't waste your time & [2123, 1005, 1056, 5949, 2115, 2051] \\
the worst movie i've seen & [5409, 3185, 1045, 2031, 2412, 2464] \\
\hline
\multicolumn{2}{l}{\textit{Positive Expressions}} \\
\hline
highly recommended & [3811, 6749] \\
must see & [2442, 2156] \\
a great deal of & [1037, 2307, 3066, 1997] \\
\hline
\multicolumn{2}{l}{\textit{Hedging Phrases}} \\
\hline
kind of, sort of & [2785, 1997], [4066, 1997] \\
in my opinion & [1999, 2026, 5448] \\
\hline
\multicolumn{2}{l}{\textit{Technical Aspects}} \\
\hline
the acting, the plot, the script  & [1996, 3772], [1996, 5436], [1996, 5896] \\
\hline
\multicolumn{2}{l}{\textit{Viewing Context}} \\
\hline
on dvd, in theaters, on screen & [2006, 4966], [1999, 4258], [2006, 3898] \\
\hline
\end{tabular}
\end{center}

\subsection{RCT - Biomedical Dataset}

Our model identifies and combines tokens related to statistical reporting, study methodology, and standard research phraseology. We present a few samples below.






\begin{center}
\small
\begin{tabular}{lc}
\hline
\textbf{Identified Pattern} & \textbf{BERT tokenizer output} \\
\hline
\multicolumn{2}{l}{\textit{P-Value Notations}} \\
\hline
p < 0.001 & [1052, 1026, 1014, 1012, 25604] \\
p = 0.05 & [1052, 1027, 1014, 1012, 5709] \\
p > 0.05 & [1052, 1028, 1014, 1012, 5709] \\
\hline
\multicolumn{2}{l}{\textit{Clinical Trial Descriptors}} \\
\hline
randomized & [6721, 3550] \\
double-blind & [3313, 1011, 6397] \\
placebo-controlled & [2173, 5092, 1011, 4758] \\
\hline
\multicolumn{2}{l}{\textit{Study Groups}} \\
\hline
control group & [2491, 2177] \\
intervention group & [8830, 2177] \\
placebo group & [2173, 5092, 2177] \\
\hline
\multicolumn{2}{l}{\textit{Study Objectives}} \\
\hline
the aim of this study & [1996, 6614, 1997, 2023, 2817] \\
to evaluate the & [2000, 16157, 1996] \\
to investigate the & [2000, 8556, 1996] \\
\hline
\multicolumn{2}{l}{\textit{Results Reporting}} \\
\hline
significant difference & [3278, 4489] \\
associated with & [3378, 2007] \\
compared with & [4102, 2007] \\
\hline
\end{tabular}
\end{center}

\section{Related Work}
\label{sec:rel_works}

Domain adaptation for pretrained models such as BERT has been extensively studied, from two independent threads: tokenization / input processing and parameter-efficient or sparse finetuning methods. Input-based approaches apply domain-specific preprocessing~\cite{iclr_token} or introduce domain-specific subword sequences to the model vocabulary, reducing the domain gap~\cite{aws, stochastic, multigrain}. On the other hand, sparse finetuning methods adapt model weights without modifying the input pipeline, including techniques like LoRA~\cite{zhang2023adalora, valipour2022dylora}, prompt tuning~\cite{lester2021power}, and selective fine-tuning~\cite{neft, leng2024towards}. AdaLoRA has been shown to outperform numerous previous adaptation methods including bias-tuning~\cite{zaken2021bitfit}, alternate adapters~\cite{houlsby2019parameter}, and heuristics that tune specific parts of the network~\cite{heuristic}, similar to our illustraion in \cref{fig:intro2}. Layer-selection has also been proposed as an alternate sparse finetuning approach~\cite{pan2024lisa, yao2024layer}. In summary, while tokenization and sparse finetuning / adaptation have been studied independently in the past, our work proposes to combine these two directions, addressing the limitations of both direct fine-tuning and adapter approaches for specialized tasks.

\label{rel:finetune}

\section{Conclusion}
\label{sec:conclusion}


In conclusion, combining sparse finetuning and token augmentation demonstrates promising results in adapting pre-trained models efficiently to specialized classification tasks. Our results across five semantic classification tasks show that task-specific token constructs, when combined with selective parameter tuning, can bridge the gap between general-purpose models and specialized classification requirements. These gains are valuable for industry applications where resource constraints and specialized domain knowledge must be balanced. Moving forward, we plan to explore mixed-precision computation (with 32-bit reserved for sensitive parameters) to further speed up training and inference. 


\section*{Limitations}



While our approach of combining sparse fine-tuning with token augmentation offers an efficient adaptation approach, we highlight a few limitations. By design, our approach tunes a small fraction of the model's parameters. For tasks that require larger deviation from the pre-trained model's knowledge base, selective parameter updating may be insufficient. Second, while token augmentation enhances the model's ability to handle domain-specific terminology, it is dependent on the quality of domain-specific terminology available in the finetuning dataset. We provide additional details about subsequence selection criteria in Appendix B, noting that focusing on commonly occurring sequences within the domain can help improve the final model. Our parameter selection depends on having a representative finetuning dataset to compute the sensitivity metric in eq. (6). If the finetuning data is inadequate or biased, the wrong parameters can be chosen, affecting the performance of the target task.
\section{Appendix}
\label{appendix:res}




\subsection{Dataset and Model Parameters}

We initialized BERT models with the official checkpoints\footnote{https://github.com/google-research/bert} and the \textit{bert-base-uncased} tokenizer. We search over the following list of parameters to find the best performing checkpoint - batch size (varied exponentially between 2 and 32), learning rate (1e-7 to 1e-4 varied exponentially), total number of training steps (varied between $10^4$ and $10^6$ exponentially), and $\beta$ in \cref{eq:sensitivity}(1e-3 to 1). We repeat the selection for the two compared baselines. We use the Phi and Gemma huggingface checkpoints - \textit{google/gemma-2-2b-it}, \textit{microsoft/Phi-3.5-mini-instruct}.


To reduce label biases and manage the overall training load, we subsample some of the larger datasets. We select the first 5k training and 1k test data points for  Hyperpartisan\footnote{https://pan.webis.de/semeval19/semeval19-web/}, from the huggingface dataset  \textit{maneshkarun/hyperpartisan-cleaned}. For RCT\footnote{https://github.com/Franck-Dernoncourt/pubmed-rct} we use the huggingface train-test split \textit{AdaptLLM/RCT}. For RT movie reviews\footnote{https://www.kaggle.com/datasets/stefanoleone992/rotten-tomatoes-movies-and-critic-reviews-dataset} we select the first 20k training and 5k test datapoints. For ACL-ARC \footnote{http://jurgens.people.si.umich.edu/citation-function/}, we correct for the overall positive-negative imbalance by selecting the first 40k positives / 100k negatives for train and  20k positives / 50k negatives for test. For SciERC\footnote{http://nlp.cs.washington.edu/sciIE/}, we use the huggingface split \textit{nsusemiehl/SciERC}. We oversampled the classes with low count so that each class had at least 800 datapoints in the train split, and we did not modify the test split.

\subsection{Task-Specific Sequence Selection}

In each dataset, we first tokenized every data point with the unmodified tokenizer. We then apply BIDE~\cite{bide} to extract sequences (min length 3, max length 20, min freq 5) over the entire dataset. This produces both contiguous and non-contiguous subsequences. We then compute their unigram-perplexities as described in \cref{alg:phrase}. We choose the threshold cutoffs so that the number of sequences is limited to 10\% of the tokenizer vocabulary. Each chosen sequence is assigned a new token-id and added to the model vocabulary. For example, consider the input (Gene BRCA1 indicates cancer predisposition, individuals with BRCA1 mutations have a higher risk). If the following sequences are chosen as new tokens - (\textbf{BRCA1}), (\textbf{cancer predisposition}) - we append the new token for \textbf{BRCA1} after the tokens for \textit{Gene} and \textit{individuals with}, and the new token for \textbf{cancer predisposition} after \textit{indicates}. 

The original tokens from the model's base tokenizer are also retained in our final tokenization. Thus, three extra tokens are added to the original tokenization of this input. To speed up token finding, we use a reverse index lookup on the original tokens (e.g., a reverse index lookup on cancer generates two candidates - cancer treatment and cancer predisposition, one of which is present in this input and will be added to the tokens).

\subsection{LLM Prompts for Evaluation}
\label{appendix:llm}

Below we provide the prompts used to test 0-shot and 10-shot LLM performance. For the 0-shot evaluations, we removed \textit{here are a few examples for reference}, and the few-shot examples. For the 10-shot evaluations, we picked 10 examples from the training corpus and ensure there is at least one example covering each class. We performed multiple runs of each dataset with different values of the temperature parameter and picked the best run for each of the three LLM models. 

\noindent \textbf{Template} - Read the following sentence closely. \underline{<\textbf{Instructions}>} + \textit{(Here are a few examples and the corresponding labels for reference - << 10 examples with labels >>)}. Please provide just your prediction for the label for the following sentence.

We specify the instructions below for each dataset.

\noindent \textbf{Hyperpartisan}: Decide whether it follows a hyperpartisan argumentation, i.e., whether it exhibits blind, prejudiced, or unreasoning allegiance to one party, faction, cause, or person. If yes, provide the label 1, if not provide the label 0.

\noindent \textbf{RCT}: Classify the sentence as one of five labels that best describes the content in relation to a research study: background, methods, results, conclusions, or objective.

\noindent \textbf{Movie Reviews}: You need to predict 0 if the review expresses a negative sentiment, and 1 if it is positive.

\noindent \textbf{ACL-ARC}: In the following sentences selected randomly from research papers, the citation tags have been deleted. You need to predict whether the sentence had a citation (label 1) or not (label 0) with just the content. If the content contains claims that require backing up, they are likely to cite a paper.

\noindent \textbf{Sci-ERC}: Please classify the relation of the text in the square bracket to the text in the pointy bracket into one of these labels - conjunction, feature-of, evaluate-for, used-for, hyponym-of, compare, part-of.

\section{GenAI Usage Disclosure}
    No generative AI tools were used to create any of the content presented in this paper. All written material in this paper and the experimental code were implemented and executed by the authors without generative AI assistance.


\bibliographystyle{ACM-Reference-Format}
\bibliography{custom}

\appendix

\end{document}